\documentclass[final,3p,times,twocolumn]{elsarticle}

\usepackage{blindtext, graphicx, amsmath, algorithm, algpseudocode, pifont, algcompatible, comment, layout, amsthm, amssymb}
\usepackage{multirow}
\usepackage{enumitem}   
\usepackage{eso-pic}
\usepackage{booktabs}
\usepackage{float}
\usepackage{color,soul}
\usepackage{xcolor}

\usepackage[utf8]{inputenc}
\usepackage[english]{babel}
\usepackage{hyperref} 
\hypersetup{ colorlinks=true, linkcolor=black, filecolor=black, urlcolor=cyan, }

\usepackage{caption}
\journal{ICT Express}

\begin{document}

\begin{frontmatter}

\title{Pre-Trained Language Models for Keyphrase Prediction: A Review}
\author{Muhammad Umair$^a$, Tangina Sultana$^{a b}$, Young-Koo Lee\corref{cor1}$^a$}

\address{$^a$ Dept. of Computer Science and Engineering, Kyung Hee University, Global Campus. Yongin-si, South Korea}

\address{$^b$Dept. of Electronics and Communication Engineering, Hajee Mohammad Danesh Science and Technology University, Bangladesh}

\cortext[cor1]{Corresponding author}
\ead{\{umair, tangina, yklee\} @khu.ac.kr}

\begin{abstract}
Keyphrase Prediction (KP) is essential for identifying keyphrases in a document that can summarize its content. However, recent
Natural Language Processing (NLP) advances have developed more efficient KP models using deep learning techniques. The limitation
of a comprehensive exploration jointly both keyphrase extraction and generation using pre-trained language models spotlights a critical gap in the literature, compelling our survey paper to bridge this deficiency and offer a unified and in-depth analysis to address limitations in previous surveys. This paper extensively examines the topic of pre-trained language models for keyphrase prediction (PLM-KP), which are trained on large text corpora via different learning (supervisor, unsupervised, semi-supervised, and self-supervised) techniques, to provide respective insights into these two types of tasks in NLP, precisely, Keyphrase Extraction (KPE) and Keyphrase Generation (KPG). We introduce appropriate taxonomies for PLM-KPE and KPG to highlight these two main tasks of NLP. Moreover, we point out some promising future directions for predicting keyphrases.
\end{abstract}


\end{frontmatter}


\section{Introduction}
To determine if a keyphrase is present in a document, it must appear as a single contiguous word. Keyphrase extraction involves using a model to accurately identify and classify the keyphrases in the document. The generation of keyphrases is another task in which the model predicts both present and absent keyphrases within the context of the document, introduced in \cite{meng-etal-2017-deep}. The application of deep learning technologies has witnessed a noticeable rise in using pre-trained language models (PLMs) in NLP in recent years. PLMs are trained using different strategies on extensive text corpora and have shown exceptional performance in various downstream tasks, including Keyphrase Predation. PLMs using self-supervised learning differ from traditional learning methods, such as supervised learning, because they are first trained on a large volume of unlabeled data before fine-tuning small quantities of labeled data for specific tasks. Self-supervised learning-based PLMs, contrary to conventional learning methods like supervised learning, are pre-trained on vast amounts of unlabeled data before being fine-tuned on tiny amounts of labeled data for particular tasks. In the realm of NLP, BERT \cite{devlin2018bert}, GPT \cite{radford2018improving}, and T5 \cite{raffel2020exploring} are some of the notable works that have consistently updated benchmark records in Pre-trained Language Model Keyphrase Extraction (PLM-KPE) and Pre-trained Language Model Keyphrase Generation (PLM-KPG) tasks \cite{liu2021addressing}, contributing significantly to the development of NLP.

\begin{table*}[]
    \centering
    \begin{tabular}{l}
    \toprule
         \textbf{\textit{Input document:}} The development of algorithms and models that allow computers to learn from data, make \\ predictions, and make decisions is known as \underline{machine learning}. This branch of \underline{artificial intelligence} involves techniques \\ like supervised, unsupervised, and reinforcement learning. \underline{Machine learning} has a vast range of applications \\ in fields such as healthcare, computer vision, and natural language processing. It is composed of feature extraction,\\ training data, and evaluation metrics, which are essential components. Deep learning, a subset of \underline{machine learning},\\ uses artificial neural networks to perform complex tasks. The advancements in \underline{machine learning} algorithms, and \\computing power has brought about significant changes in different industries, leading to breakthroughs in areas\\ like fraud detection, autonomous vehicles, and recommendation systems. \\
         \midrule
         \textbf{\textit{Keyphrase Extraction:}} \underline{Machine learning}, \underline{artificial intelligence}, Algorithms, Models, Data \\
         \midrule
         \textbf{\textit{Keyphrase Generation:}} Self-guided Learning, Natural language processing, Reward-based Learning,\\ Textual Understanding \\
         \bottomrule
    \end{tabular}
    \caption{An illustration of the Top 5 keyphrase extraction and generation process. The keyphrases currently featured in the document have been underlined for your convenience.}
    \label{tab:example}
\end{table*}

The process of extracting keyphrases from a document involves identifying and extracting significant phrases that represent the main topics or concepts discussed within it. The primary objective is to extract the most essential and representative phrases using feature-based \cite{campos2018yake, hulth2003improved, ohsawa1998keygraph, turney2002learning, zhang2006keyword} and linguistic techniques \cite{el2012arabic} like frequency analysis \cite{salton1988term}, part-of-speech tagging \cite{barker2000using, Mihalcea2004404}, and syntactic parsing \cite{le2016unsupervised}. These methods can identify keyphrases based on their frequency, relevance, or structural patterns within the text, allowing for a more thorough analysis and understanding of the document's contents. On the other hand, the keyphrase generation involves creating new phrases that are not present in the original document. The objective is to generate concise and meaningful phrases that encapsulate the central concepts or themes discussed in the document. Keyphrase generation methods rely on language modeling \cite{lewis2019bart, raffel2020exploring, kulkarni2021learning}, neural networks \cite{sharkey1996combining}, or rule-based systems \cite{karad2015rule} to generate coherent and significant keyphrases.

Table \ref{tab:example} provides an illustrative example of the keyphrase prediction process, which involves detecting two distinct categories of keyphrases in a given document. The first category comprises keyphrases that are consistently present throughout the document, while the second category encompasses those that cannot be found in any contiguous subsequence of the document. In the past, researchers primarily focused on keyphrase extraction, which sought to extract keyphrases directly from the document to improve keyphrase prediction. However, with the emergence of deep learning, researchers are currently exploring keyphrase production using dominant models that can generate both present and absent keyphrases.

The limitations in the existing literature become evident as we observe the absence of a comprehensive exploration that integrates both keyphrase extraction and generation using pre-trained language models, leaving a critical gap in understanding the unified process of keyphrase prediction. Previous studies have offered fragmented insights, often focusing solely on extraction or generation, thus limiting a holistic understanding of the field. Additionally, the need for standardized taxonomies further compounds the issue, creating confusion and hindering the systematic exploration of keyphrase prediction methods. These limitations emphasize the necessity for our survey paper, which aims to bridge these gaps by providing a unified and comprehensive analysis, introducing clear taxonomies, and exploring promising future directions to advance the understanding and application of keyphrase prediction in Natural Language Processing.

\begin{figure*}[!h]
  \centering
  \includegraphics[width=\linewidth]{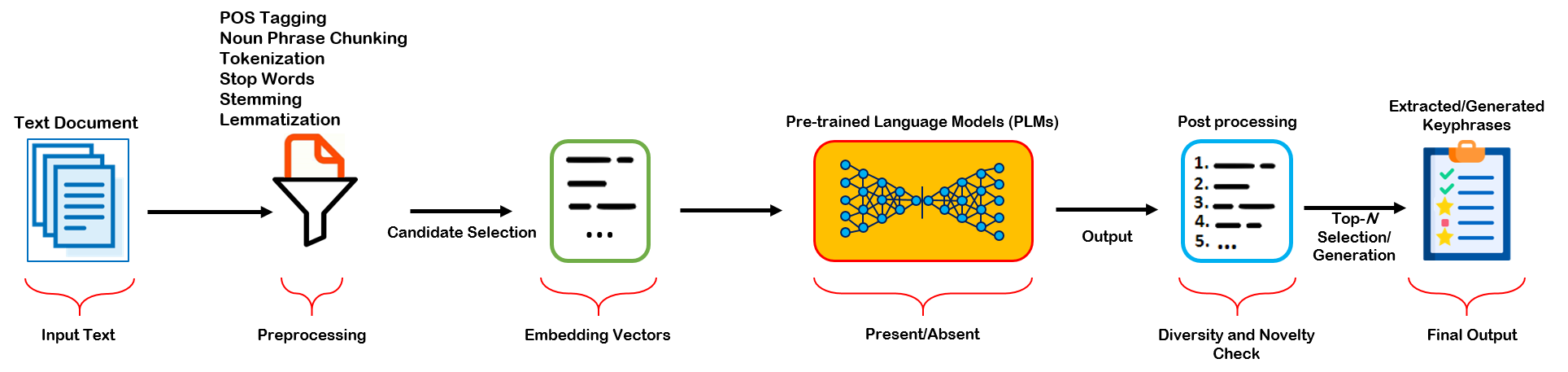}
  \caption{Overview of Keyphrase Prediction using Pre-trained Language Models for Keyphrase Extraction and Keyphrase Generation.}
  \label{KPE_Overview}
\end{figure*}

The objective of this paper is to furnish a succinct synopsis of the current methodologies employed for keyphrase extraction and generation, focusing on the utilization of pre-trained language models. To avoid ethical concerns, we avoid proposing new models and instead attempt to enhance the research community's understanding in this domain. The main emphasis of this research is on the latest developments in predicting keyphrases using neural networks into language models that can promote both PLM-KPE and PLM-KPG tasks, and these two areas have different focuses. We aim to present a comprehensive review of Pre-trained Language Models Keyphrase prediction (PLM-KP) in the two areas to provide respective insights into PLM-KP in PLM-KPE and PLM-KPG. The survey's primary contributions can be summarized as follows.

\begin{enumerate} 
\item Our survey paper addresses the gaps in the literature by providing a unified and comprehensive analysis that bridges the divide between keyphrase extraction and generation. This approach enhances the overall understanding of keyphrase prediction in the context of Natural Language Processing.

\item We introduce structured taxonomies for both PLMs for keyphrase prediction (PLM-KP) and its constituent tasks (KPE, KPG, and Multimodal) in Figure. \ref{main_taxonomy} shows our proposed taxonomies. These taxonomies offer a clear and systematic framework, facilitating a cohesive classification of methods and techniques.

\item For PLM-KPE, PLMs are further divided into four subcategories according to the types of methods: Attention Mechanism, Graph-based Ranking, Graph-based Ranking, and Phrase-Document Similarity.

\item For PLM-KPG, we focus on the domain specific and low-resources methods.

\item In addition to addressing existing gaps, our paper identifies and discusses promising future directions in keyphrase prediction. By pointing out these areas, we guide future research endeavors and stimulate innovation in keyphrase prediction.
\end{enumerate}

The remainder of this paper is structured as follows. The section \ref{pre} clearly defines the previous survey and the uniqueness of our work, followed by the research criteria. Next, The background of PLM-KP in the context of NLP training paradigms is discussed in Section \ref{background}, followed by an explanation of the research methodology utilized for this analysis. In Section \ref{kpe}, we introduce the taxonomy of PLM-KP in the field of KPE and provide a comprehensive overview of the prevalent methods used in keyphrase extraction. Section \ref{multi} introduces the Multimodel used in keyphrase extraction and generation. In Section \ref{kpg}, we introduce the taxonomy of PLM-KP in the field of KPG. For both KPE and KPG fields, we focus on discussing the exemplary works within each category of leaves in the taxonomy. Next, we compare the models in \ref{modelcomparison}. Section \ref{dataset} shows the publicly available dataset for keyphrase prediction and the PLM-KP SOTA techniques. In Section \ref{AppChalFRD}, we explore the diverse applications in which pre-trained language models highlight unique aspects, including Natural Language Understanding (NLU), Text Generation, Translation and Language Modeling, Information Retrieval and Summarization, Health and Medicine, Ethical and Societal Applications. We thoroughly analyze the current limitations and challenges of PLM-KP in section \ref{challenges}, providing insights. We have identified potential areas for future research with the novel contribution of this work, which is referenced in \ref{future}. In Section \ref{concl}, we provide our concluding remarks.

\section{Preliminary}
\label{pre}
This section outlines the distinctions between our current survey and those conducted previously. Following this comparison, we will outline the specific research criteria that guided the formulation and focus of our paper.

\subsection{Previous Surveys}
The comprehensive review of keyphrase extraction \cite{https://doi.org/10.1002/widm.1339} highlights KPE methods, categorizing them into unsupervised and supervised approaches. In contrast, our paper includes more recent studies published explicitly in keyphrase generation tasks and multimodal keyphrase extraction and generation.

Advancements in pre-trained language models have significantly contributed to both keyphrase extraction and generation tasks. Unlike a recent keyphrase extraction survey \cite{song2023survey}, which mainly includes supervised and unsupervised keyphrase extraction in two stages without covering keyphrase generation, our paper is a comprehensive resource for both novices and experts. We have consolidated information on both KPE and KPG in one location. Furthermore, our paper reviews 22 more recent articles than \cite{xie2023statistical}, ensuring it is up-to-date. \cite{JournalArticle, aydin2020review, sharma2021keyphrase, alami2020automatic, ajallouda2023automatic, glazkova2023applying, ajallouda2022systematic} Unlike previous surveys, our work uniquely summarizes SOTA attention models in Section \ref{am}, aiming to enhance the research community's understanding in this domain. This comprehensive overview serves as a pivotal resource, providing insights that are critical for advancing keyphrase extraction and generation research.

Existing surveys on keyphrase prediction primarily focus on early feature-engineered and neural-based keyphrase extraction models. However, no current and comprehensive survey provides detailed knowledge of KPE and KPG tasks that leverage pre-trained language models. Our work fills this gap by thoroughly reviewing the latest advancements in the field.

\begin{table}[!h]
    \centering
    \begin{tabular}{ll}
    \toprule
       \# & online Database \\
       \midrule
        OBD1 & Google Scholar \\
        ODB2 & ACM Digital Library \\
        ODB3 & Science Direct \\
        ODB4 & SpringerLink \\
        ODB5 & IEEE Xplore \\
        ODB6 & Cornell Arxiv (cs) \\
       \bottomrule 
    \end{tabular}
    \caption{List of the online databases examined in this survey.}
    \label{tab:onlineDBs}
\end{table}

\subsection{Research criteria}
Efficient and reliable methods for gathering relevant literature are crucial when conducting a survey. Our approach involves utilizing query-based criteria across a range of reputable online databases, which are listed in Table \ref{tab:onlineDBs}. Additionally, we meticulously examine respected journals, conferences, and workshops to analyze work related to keyphrases from various perspectives in a comprehensive way.

The survey selection process comprises three essential steps for databases and targeted journals/conferences. Initially, we use primary keywords, such as "keyphrase extraction," "keyphrase generation," and "pre-trained language models based keyphrase," to conduct formal searches sequentially across the chosen databases and target domain journals and conferences. Potential works are then critically evaluated based on predetermined criteria for inclusion. Lastly, we carefully consider each paper's titles, abstracts, and full texts during each formal search. As demonstrated in Table \ref{tab:topConf}, a significant amount of academic literature is available on keyphrase extraction and generation, with a strong presence at leading computer science conferences. This finding suggests continued interest and ongoing research in this field of study over an extended period.

\section{Background and Summaries of LLMs}
\label{background}
The surge in PLM advancements in NLP has significantly influenced keyphrase prediction (KP) methodologies. This section elucidates the comprehensive journey from raw text input through various processing stages, leveraging PLMs for KP, as depicted in Figure \ref{KPE_Overview}. Moreover, this section sheds light on several state-of-the-art PLMs, including BERT, Roberta, ELMo, Llama 2, T5, and GPT. It provides comparative analysis and summaries of their functionalities, strengths, and potential in KP tasks. By integrating these PLMs, significant strides have been made in both KPE and KPG, showcasing their indispensable role in enhancing NLP tasks.

\subsection{Background}
The following section will dive into a step-by-step breakdown of Figure \ref{KPE_Overview}, expounding each component and process involved in keyphrase extraction and generation using pre-trained language models.

\begin{table*}[ht]
\centering
\begin{tabular}{p{2.5cm}p{3.5cm}p{3.5cm}p{3.5cm}}
\toprule
\textbf{Model} & \textbf{Key Features} & \textbf{Strengths for KP} & \textbf{Limitations} \\
\midrule
\textbf{BERT} & Bidirectional training, Transformer-based & Deep context understanding, adaptable for KP tasks & High computational demand \\
\midrule
\textbf{RoBERTa} & Enhanced BERT with dynamic masking and extended training & Improved accuracy, better context sensitivity for KP & Higher computational cost than BERT \\
\midrule
\textbf{ELMo} & Contextualized word representations & Nuanced meaning capture for precise KP & High computation for large docs \\
\midrule
\textbf{Llama2} & Vast and diverse training datasets & Scalable and adaptable for various KP tasks & Less tested in practice compared to others \\
\midrule
\textbf{T5} & Text-to-text framework, versatile NLP tasks applicability & Models KP as a natural language task, easily fine-tuned & Requires task-specific tuning \\
\midrule
\textbf{GPT} & Advanced autoregressive language model, large-scale training & High generative capacity, excellent at generating contextually relevant keyphrases & Generates verbose outputs, may require post-processing \\
\bottomrule
\end{tabular}
\caption{Comparative overview of six advanced language models, including BERT, Roberta, ELMo, Llama, T5, and GPT. This analysis highlights each model's core features, their applicability to keyphrase prediction (KP) tasks, and their inherent limitations, offering a comprehensive view of their contributions to enhancing KP methodologies.}
\label{tab:model_comparison}
\end{table*}

\begin{itemize}
\item Step 1: Input from Documents
in the initial step, raw textual data from documents serves as input. This unprocessed text provides the foundational material from which keyphrases will be extracted or generated.
\item Step 2: Processing (POS Tagging, Noun Phrase Chunking, Tokenization, Stop Word Removal, Stemming, and Lemmatization)
In this phase, the input text undergoes a series of preprocessing steps to enhance its structure and facilitate meaningful analysis. These steps include Part-of-Speech (POS) tagging \cite{schopf2022patternrank, bennani2018simple, ding2021attentionrank, sun2020sifrank}, which identifies grammatical categories of words; noun phrase chunking \cite{sun2020sifrank, wu2021unikeyphrase, giarelis2023lmrank}, isolating noun phrases for analysis; tokenization \cite{kim2021keyword, liu2020keyphrase, liang2021unsupervised}, breaking down the text into individual tokens; removal of stop words \cite{cheng2023w2kpe, popova2018keyphrase}, eliminating common words with limited semantic meaning \cite{zahera2022multpax}; stemming \cite{ding2022agrank, 10.1007/978-3-031-45275-8_10, safari2022classification}, reducing words to their root form; and lemmatization \cite{alharbi2022arabic, safari2022classification, gagliardi2020semantic}, reducing words to their base or dictionary form. Each of these processes refines the text, preparing it for further analysis.
\item Step 3: Embedding (Vector representation) Following pre-processing, the processed text is transformed into vector representations \cite{xiong2019open, zhu2020deep, zhou2021topic}. This embedding step converts words or phrases into numerical vectors, capturing their semantic meanings in a multidimensional space. Embeddings facilitate the modeling of word relationships and context, enabling the pre-trained language models to grasp the nuances of the text.
\item Step 4: Pre-Trained Language Models (Extraction or Generation)
In this pivotal phase, the processed and embedded text is fed into pre-trained language models. These sophisticated models, often based on transformer architectures, can understand complex linguistic patterns. The models identify and extract relevant phrases from the input text for keyphrase extraction. Alternatively, for keyphrase generation, the models create new, contextually relevant keyphrases based on the learned patterns and embeddings.
\item Step 5: Post-Processing (Diversity and Novelty Check)
After the keyphrases are extracted or generated, a post-processing \cite{wang2014corpus} step ensues. Here, the extracted keyphrases undergo a diversity \cite{carbonell1998use, sun2021capturing, sun2019divgraphpointer, sun2020sifrank} and novelty check. This ensures that the selected keyphrases are pertinent to the document's context and unique and varied \cite{bennani2018simple}. Post-processing \cite{wang2014corpus, devika2021deep, duari2020complex} is crucial in refining the final selection of keyphrases, enhancing their relevance and richness.
\item Step 6: Extracted or Generated Keyphrases
The culminating step presents the output, the extracted or generated keyphrases. These keyphrases, refined through careful preprocessing, embedding, modeling, and postprocessing, capture the essence of the document. They serve as concise and meaningful representations of the document's content, facilitating efficient information retrieval and understanding.
\end{itemize}

\begin{table*}[]
    \centering
    \begin{tabular}{lcccccc}
    \toprule
        \textbf{Conferences} & \textbf{2018} & \textbf{2019} & \textbf{2020} & \textbf{2021} & \textbf{2022} & \textbf{2023}\\
         \midrule
         \textbf{Keyphrase extraction} \\
         \midrule
         AAAI & 0 & 0 & 0 & 0 & 0 & 0 \\
         COLING & 0 & - & 4 & - & 0 & - \\
         NAACL & 1 & 1 & - & 2 & 1 & - \\
         EMNLP & 0 & 1 & 1 & 3 & 0 & 4 \\
         ACL & 0 & 1 & 0 & 0 & 0 & 3 \\
         \midrule
        \textbf{Keyphrase generation} \\
         \midrule
         AAAI & 0 & 1 & 0 & 1 & 2 & 0 \\
         COLING & 0 & - & 1 & - & 1 & - \\
         NAACL & 0 & 2 & - & 3 & 2 & - \\
         EMNLP & 2 & 0 & 3 & 3 & 5 & 2 \\
         ACL & 0 & 3 & 3 & 2 & 0 & 3 \\
         \midrule
         \textbf{Total} & 3 & 9 & 12 & 14 & 11 & 12 \\
        \bottomrule     
    \end{tabular}
    \caption{Papers on keyphrase extraction and generation have been presented at major computer science conferences. Conferences marked with '-' are either not being held or scheduled for the future.}
    \label{tab:topConf}
\end{table*}

\subsection{Summaries of Large Language Models}
To better understand the advancements in language models and their impact on keyphrase prediction, we have compiled a comparative analysis (see Table \ref{tab:model_comparison}). This analysis delineates six pivotal models' unique features, strengths, and limitations. BERT \cite{devlin2018bert}, RoBERTa \cite{liu2019roberta}, ELMo \cite{peters2018deep}, Llama2 \cite{touvron2023llama}, T5 \cite{raffel2020exploring}, and GPT \cite{openai2024gpt4}. Each model brings distinct advantages to the KP domain. Moreover, pre-trained language models \cite{brown2020language, lewis2019bart, koubaa2023gpt, reimers2019sentence, yang2019xlnet, lan2019albert} have significantly advanced the current state of NLP tasks. As a result, they are now being integrated into methods for both KPE and KPG, such as the notable examples are the use of PLM-KP for unsupervised KPE \cite{sun2020sifrank, liang2021unsupervised}. In this approach, KPE is achieved through sequence labeling, \cite{sahrawat2019keyphrase, dascalu2021experiments}, while KPG utilizes the sequence-to-sequence technique \cite{9443960, 9576585, chowdhury2022applying, kulkarni-etal-2022-learning, gao-etal-2022-retrieval, wu2022representation}. Notable advancements have been made by models such as ELMo \cite{peters2018deep}, and BERT \cite{devlin2018bert}, which are based on LSTM and Transformer architectures \cite{vaswani2017attention}, respectively. Transformer-based models, in particular, utilize a masked language model and sentence adjacency training objectives to learn bidirectional representations of words.

\begin{figure*}[!h]
  \centering
  \includegraphics[width=\linewidth]{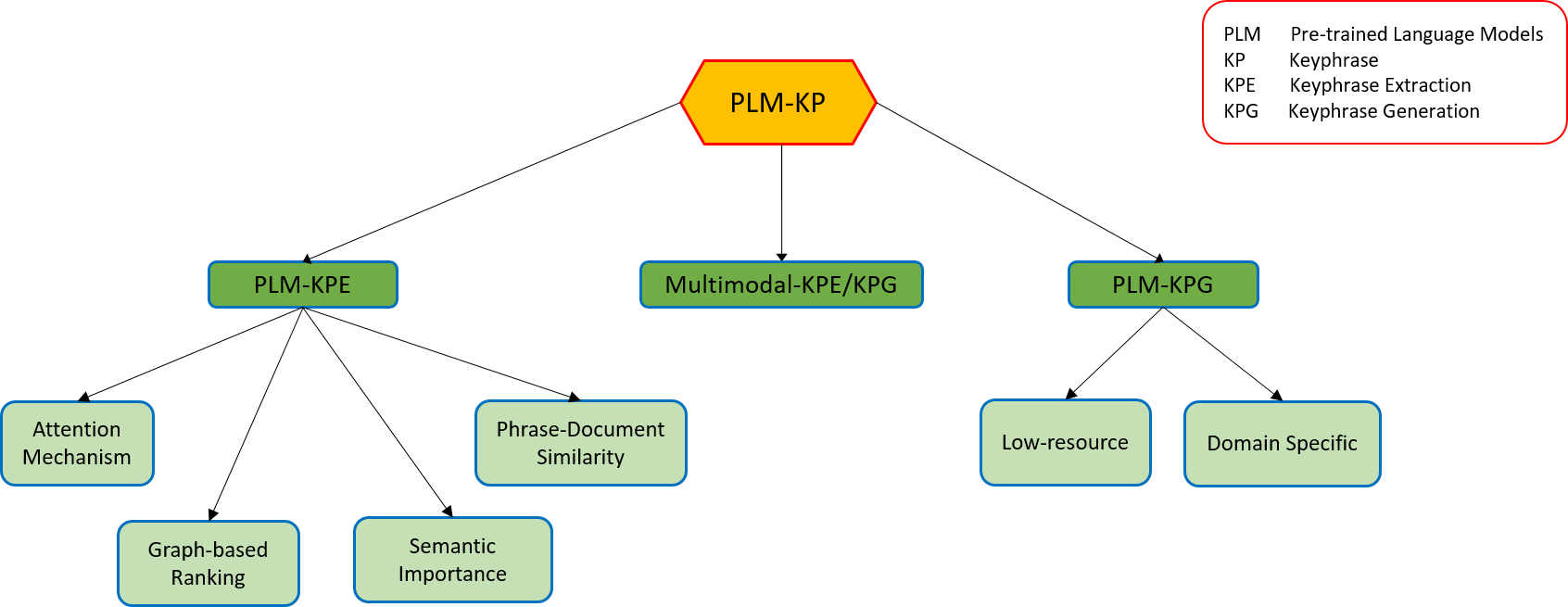}
  \caption{Categorization of Keyphrase Prediction using Pre-trained Language Models (PLM-KP), centered around two fundamental NLP tasks: Pre-trained Language Model Keyphrase Extraction (PLM-KPE) and Pre-trained Language Model Keyphrase Generation (PLM-KPG).}
  \label{main_taxonomy}
\end{figure*}

\begin{itemize}
\item BART (Bidirectional and Auto-Regressive Transformers) leverages a novel pre-training objective of text infilling, where a certain percentage of the input text is masked, and the model learns to predict the masked portions. This approach, combined with its bidirectional context understanding, makes BART highly effective for KP tasks, as it can generate coherent and contextually relevant keyphrases by comprehensively understanding the given text's nuances.

\item RoBERTa (Robustly Optimized BERT Approach) builds upon the original BERT framework through optimized training strategies, including dynamic masking and extended training over more data. This enhancement enables Roberta to outperform its predecessor in understanding and generating text, making it highly efficient for KP tasks. RoBERTa's improved contextual embeddings contribute significantly to extracting semantically rich and relevant keyphrases from complex documents, demonstrating its prowess in capturing the essence of textual content.

\item ELMo (Embeddings from Language Models) introduces the concept of deep contextualized word representations, where the meaning of each word can change based on the surrounding text. This feature is particularly beneficial for KP, as it allows for extracting keyphrases that are contextually aligned with the document's overall theme. ELMo's ability to account for word-level nuances dramatically enhances the precision of keyphrase identification, making it a valuable asset for tasks requiring deep semantic understanding.

\item Llama2 (Language Model from Meta AI), though relatively new, has shown promising capabilities in understanding and generating natural language text. Its training on diverse datasets allows for a broad comprehension of language nuances, making it adept at generating coherent and context-relevant keyphrases. Llama's versatility and scalability position it as an emerging tool for KP tasks, potentially offering novel approaches to keyphrase extraction and generation.

\item T5 (Text-to-Text Transfer Transformer) adopts a unified framework that converts all NLP problems into a text-to-text format, whether translation, summarization, or keyphrase generation. This simplification allows T5 to leverage its vast pre-trained knowledge, fine-tuned with task-specific data, to produce highly relevant and accurate keyphrases. Its capacity to understand and generate text based on the provided prompts makes it particularly useful for both extraction and generation tasks within KP.
\end{itemize}

\begin{table*}[t]
    \centering
    \begin{tabular}{p{0.2\linewidth}p{0.35\linewidth}p{0.35\linewidth}}
    \toprule
      \textbf{Technique} & \textbf{Advantages} & \textbf{Disadvantages} \\
      \midrule
       Attention Mechanism & - Captures contextual relevance.\newline - Dynamically weights inputs.\newline - Enhances interpretability. & - High computational cost.\newline - Needs large training data. \\
       \midrule
       Graph-based Ranking & - No labeled data needed.\newline - Fits unsupervised learning.\newline - Scalable. & - Quality of graph is crucial.\newline - Might miss nuanced meanings. \\
       \midrule
       Semantic Importance & - Finds semantically valuable keyphrases.\newline - Uses semantic networks/embeddings.\newline - Effective across domains. & - Complex semantic analysis needed.\newline - Struggles with novel terms. \\
       \midrule
       Phrase-Documents Similarity & - Measures content relevance directly.\newline - Identifies representative phrases.\newline - Suitable for supervised/unsupervised settings. & - Limited contextual capture.\newline - Needs to be tuned similarity measures. \\
    \bottomrule
    \end{tabular}
    \caption{Comparative Analysis of Keyphrase Extraction Techniques: Advantages and Disadvantages.}
    \label{tab:ComparativeAnalysis}
\end{table*}

\section{PLMs for KPE}
\label{kpe}
To provide a comprehensive overview of the prevalent methods used in keyphrase extraction, Table \ref{tab:ComparativeAnalysis} presents a comparative analysis focusing on the advantages and disadvantages of each technique. This analysis emphasizes the importance of selecting an appropriate keyphrase extraction method based on the specific requirements and constraints of the task.

Categorization of Keyphrase Prediction using Pre-trained Language Models (PLM-KP), centered around two fundamental NLP tasks: Pre-trained Language Model Keyphrase Extraction (PLM-KPE) and Pre-trained Language Model Keyphrase Generation (PLM-KPG) is shown in Figure \ref{main_taxonomy}.

\subsection{Attention Mechanism}
\label{sub_section_AM}
Extensive research has demonstrated that attention mechanisms for keyphrase extraction using pre-trained language models have made significant contributions. UCPhrase \cite{10.1145/3447548.3467397} a novel approach for identifying high-quality phrases using silver labels derived from word sequences that frequently co-occur within documents. This method could significantly improve the accuracy and efficiency of phrase tagging. These contextually rooted silver labels enhance the preservation of contextual completeness and the capture of emerging domain-specific phrases. In contrast, AttentionRank \cite{ding-luo-2021-attentionrank} presents a hybrid attention model that uses self-attention and cross-attention mechanisms to assesImportancertance of the candidate and establish semantic relevance between candidates and document sentences. While UCPhrase excels at capturing quality phrases, AttentionRank showcases robust performance across various document lengths. Integrating accumulated self-attention and cross-attention within AttentionRank paves the way for further investigations, such as domain-specific fine-tuning and comparisons against baselines on specific datasets.

\subsection{Graph-based Ranking}
\label{sub_section_GBR}
Pioneering work has incorporated pre-trained language models into graph-based keyphrase extraction techniques, significantly enhancing the extraction process. Mahata et al. \cite{mahata-etal-2018-key2vec} introduces an innovative approach that uses phrase embeddings to rank keyphrases extracted from scientific articles, employing a theme-weighted PageRank methodology to achieve state-of-the-art results. Liang et al. \cite{liang2021keyphrase} transcends traditional co-occurrence models by infusing semantic information into the graph structure using word embeddings, improving benchmark datasets' performance. Rafiei et al. \cite{asl2020gleake} takes an integrative approach, using single- and multiword embeddings to construct embedding-based graphs, effectively capturing local and global context for high-quality keyphrase extraction in diverse textual domains. Yuchen et al. \cite{liang-etal-2021-unsupervised} introduce a unique boundary-aware centrality method to enhance local information capture within graph-based models, showcasing the power of pre-trained embeddings in refining both local and global context for superior keyphrase extraction performance. Together, these works underscore the transformative potential of pre-trained language models in advancing the effectiveness of graph-based keyphrase extraction techniques.

\subsection{Semantic Importance}
\label{sub_section_SI}
Semantic Importance, a pivotal aspect of keyphrase extraction, is harnessed by innovative research efforts that leverage pre-trained language models to uncover keyphrases' true significance and relevance in unsupervised settings. More recent attention has focused on the provision of semantic Importance. These two ground-breaking studies show how to use pre-trained language models to capture the semantic significance of keyphrases in unsupervised keyphrase extraction. Zhang et al. \cite{zhang2023mderank} The unique MDERank method, which employs a masking strategy and ranks potential keyphrases based on the similarity between the embeddings of the source material and the masked document, solves the shortcomings of existing techniques in dealing with lengthy documents. The paper also introduces KPEBERT, a keyphrase-oriented BERT model, and showcases how MDERank benefits from it, resulting in significant performance improvements.
On the other hand, Rishabh et al. \cite{joshi2023unsupervised} introduce INSPECT, a unique methodology that uses self-explaining models to identify influential keyphrases by measuring their predictive impact on downstream tasks, such as document topic classification. INSPECT establishes itself as a state-of-the-art solution by avoiding heuristic importance, outperforming previous approaches across various datasets. These papers collectively emphasize the potential of pre-trained language models to enhance keyphrase extraction by capturing semantic relevancy and interpretability.
\cite{wu2023kpeval} a new evaluation framework named KPEVAL, designed to assess the performance of keyphrase extraction systems more comprehensively than traditional exact match metrics. KPEVAL conducts semantic-based evaluation on reference agreement, faithfulness, diversity, and utility of keyphrase systems, providing a more comprehensive assessment compared to exact matching methods. The strong performance of LLMs, particularly GPT-3.5, in keyphrase prediction tasks encourages further exploration and improvement of Language Model Methods (LLMs) for keyphrase prediction.

\subsection{Phrase-Documents Similarity}
\label{sub_section_PDS}
Phrase-document similarity, a critical factor in keyphrase extraction, serves as the foundation for various techniques that aim to accurately identify and rank keyphrases within text documents using the power of pre-trained language models. The significance of each potential phrase is typically determined by how well the phrase matches the representation of the document. \cite{bennanismires2018simple} EmbedRank takes advantage of Sent2Vec \cite{pagliardini-etal-2018-unsupervised} and Doc2Vec \cite{le2014distributed} techniques to transform candidates and input documents into vectors. In the pursuit of diverse keyphrases, EmbedRank+ delves into the interplay of candidate similarities. In contrast, SIFRank \cite{8954611} crafts vector representations for candidates, sentences, and input documents using weighted averages, using ELMo embeddings \cite{peters2018deep}. Adding a layer of complexity, SIFRank+ considers candidate positions within the document. Notably, \cite{li2021unsupervised} enhances SIFRank's prowess by amalgamating domain relevance and phrase quality into the ranking framework. On a different note, \cite{PAPAGIANNOPOULOU2018888} approach the challenge by harnessing entire documents to train Glove embeddings \cite{pennington-etal-2014-glove}, subsequently assessing candidates based on cumulative word-document similarities.

\section{Multimodal Keyphrase Extration and Generaion}
\label{multi}
The advent of multimodal data, which includes combinations of text, image, video, and audio, presents both challenges and opportunities for keyphrase extraction and generation. In recent years, the field of Natural Language Processing (NLP) has witnessed a growing interest in leveraging these diverse data types to enhance the understanding and generation of key textual elements like keyphrases. Multimodal keyphrase extraction and generation extend beyond traditional text-based approaches by incorporating additional modalities, thus aiming to achieve a more holistic understanding of content. This integration allows for the extraction and generation of keyphrases that are not only contextually richer but also more aligned with the nuanced interplay of visual and textual cues.

Significant advancements in multimodal keyphrase prediction are highlighted by pioneering models and methodologies that effectively harness the synergy of textual and visual information to refine the accuracy and applicability of keyphrase annotations. For instance, the integration of visual entity enhancement and multi-granularity image noise filtering has proven effective. These techniques improve the semantic alignment between visual and textual elements and ensure that only relevant visual data is considered during keyphrase generation. Rigorous experimental validations on benchmark datasets have demonstrated superior performance compared to existing methods, thereby confirming the effectiveness of these visual enhancements \cite{10.1145/3581783.3612413}.

Moreover, the introduction of the One2MultiSeq training paradigm and the CopyBART model marks another innovative approach, particularly in handling social media content. This methodology not only balances the model's attention across various keyphrase types but also significantly improves the ability to generate 'absent' keyphrases often overlooked by conventional methods. By dynamically adjusting to varied keyphrase orders and integrating textual and visual content, this model addresses the challenges posed by the noisy, real-world data of social media platforms \cite{yu2024training}.

Additionally, frameworks like SMART-KPE (Strategy-based Multimodal Architecture for KeyPhrase Extraction) utilize both micro-level visual features, such as font size and color, and macro-level features like webpage layout. This approach not only enhances keyphrase prediction but also provides a deeper understanding of content, making it particularly effective in the diverse and visually rich environment of open-domain web pages \cite{wang2020incorporating}.

Furthermore, incorporating cognitive signals derived from EEG and eye-tracking into Automatic Keyphrase Extraction (AKE) offers a novel approach to extract key information from microblogs. This multimodal strategy combines deep cognitive insights and surface-level attention details to significantly enhance AKE's accuracy, especially useful for the unstructured and vast content typical of social media \cite{yan2024utilizing}.

Finally, the unified framework utilizing Multi-Modality Multi-Head Attention (M3H-Att) in social media posts leverages both text and image data to capture the complex interactions between these inputs. By integrating OCR text and image attributes, this method not only boosts keyphrase prediction accuracy but also enriches the understanding of multimodal interactions, essential for processing social media content effectively \cite{wang2020cross}.

These diverse approaches reflect the dynamic evolution of keyphrase extraction and generation methodologies, demonstrating the potential of multimodal data to enrich NLP applications. By integrating various data types, researchers and practitioners can achieve a more nuanced and comprehensive understanding of content, paving the way for more sophisticated and contextually aware NLP systems.

\section{PLMs for KPG}
\label{kpg}
\subsection{Low-resource}
\label{sub_section_Low}
\begin{figure}[h]
  \centering
  \includegraphics[scale=0.8]{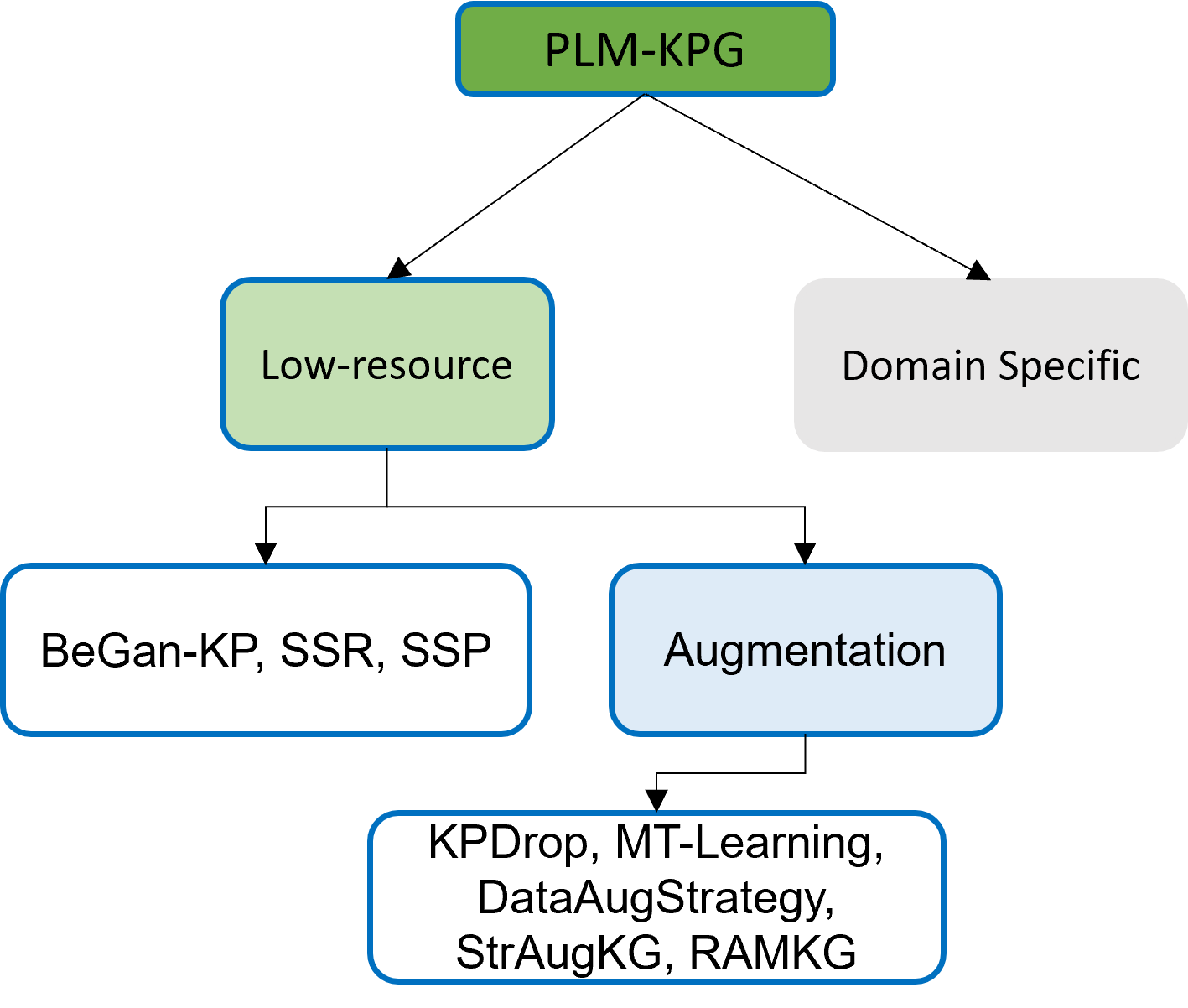}
  \caption{Categorization of Low-resource}
  \label{Low}
\end{figure}

Categorization of low-resource is shown in Figure \ref{Low}.

In the field of low-resource keyphrase generation, conventional approaches have typically relied on the application of semi-supervised or unsupervised learning techniques \cite{ray-chowdhury-etal-2022-kpdrop, ye-wang-2018-semi, wu-etal-2022-representation}. The amount and quality of available training data determines the effectiveness of keyphrase generation models. However, commonly used labeled datasets are often limited in size, making low-resource keyphrase generation an important and worthwhile research area. In this subsection, we will focus on the use of pre-trained language models for keyphrase generation in low-resource settings. A great deal of previous research \cite{Liu2020KeyphrasePW, chowdhury2021kpdrop, lancioni-etal-2020-keyphrase, garg2023data, garbacea2022adapting, wu2022representation, wu2022pre, kim-etal-2021-structure, gao-etal-2022-retrieval, Wu_2022, piedboeuf2024data}, has shown that keyphrase generation is a challenging task in low-resource settings.

In the realm of Keyphrase Generation using Pre-trained Language Models, recent advancements have illuminated multiple strategies to tackle the challenges of low-resource scenarios. In particular, Lancioni et al. \cite{lancioni2020keyphrase} introduces BeGan-KP, a GAN model for the generation of low-resource keyphrases. It features a BERT-based discriminator architecture that efficiently distinguishes between human-curated and generated keyphrases. BeGan-KP demonstrates competitive results on multiple datasets with less than 1\% of the training data, making it effective in low-resource scenarios. Meanwhile, Di Wu et al. \cite{wu2022representation} presents a data-oriented approach for low-resource keyphrase generation. Leveraging retrieval-based statistics and pre-trained language models, it learns intermediate representations for improved performance. The proposed approach facilitates the generation of absent keyphrases and shows promising results even with limited training examples.

Research in this area has shown that data augmentation plays a critical role in low-resource settings for keyphrase generation. In particular, Ray et al. \cite{chowdhury2021kpdrop} introduces KPDROP, a model-agnostic approach to improve the generation of absent keyphrases. It achieves this by randomly dropping present keyphrases during training, encouraging the model to better infer relevant but absent keyphrases. KPDROP proves effective in improving both present and absent keyphrase generation performance in various settings. Expanding further, Cristina et al. \cite{garbacea2022adapting} explores the adaptation of pre-trained language models for low-resource text simplification. It compares meta-learning and fine-tuning approaches and finds that structured intermediate adaptation steps lead to significant performance gains. This work paves the way for more comprehensive adaptive learning solutions. Building on this momentum, Krishna et al. \cite{garg2023data} focuses on data augmentation strategies tailored for low-resource keyphrase generation. The methods utilize full-text articles to enhance the generation of present and absent keyphrases. These strategies consistently outperform existing approaches, demonstrating their effectiveness in resource-constrained scenarios. Complementing this, Jihyuk kim et al. \cite{kim2021structure} addresses keyphrase generation in scenarios where structure plays a pivotal role. The approach augments documents with related keyphrases and encodes structure-aware representations using graphs. This strategy significantly improves keyphrase generation for documents with varying structures. Extending the scope, Yifan et al. \cite{gao2022retrieval} tackles multilingual keyphrase generation using a retrieval-augmented approach. By leveraging English keyphrase annotations and cross-lingual retrieval, the model generates keyphrases in low-resource languages. The proposed iterative training algorithm for retrievers and generators further enhances cross-lingual retrieval. This method outperforms baselines and offers promising results in multilingual settings. These approaches collectively pave the way for innovative methods in the realm of keyphrase extraction and generation using pre-trained language models. \cite{wu2023rethinking} another SOTA technique can be presented under the umbrella of data augmentation. It introduces significant advancements through its DESEL (Decode-Select) algorithm, which enhances keyphrase generation by integrating the precision of greedy search with the recall benefits of sampling methods. DESEL first decodes a sequence using greedy search, then samples additional sequences to create a pool of candidate keyphrases, and selects the most probable phrases to improve overall prediction quality. This method effectively balances precision and recall, leading to superior performance across multiple datasets, demonstrating a refined approach to using pre-trained language models for keyphrase generation.

\subsection{Domain-specific}
\label{sub_section_DM}

\begin{figure*}[h]
  \centering
  \includegraphics[width=\linewidth]{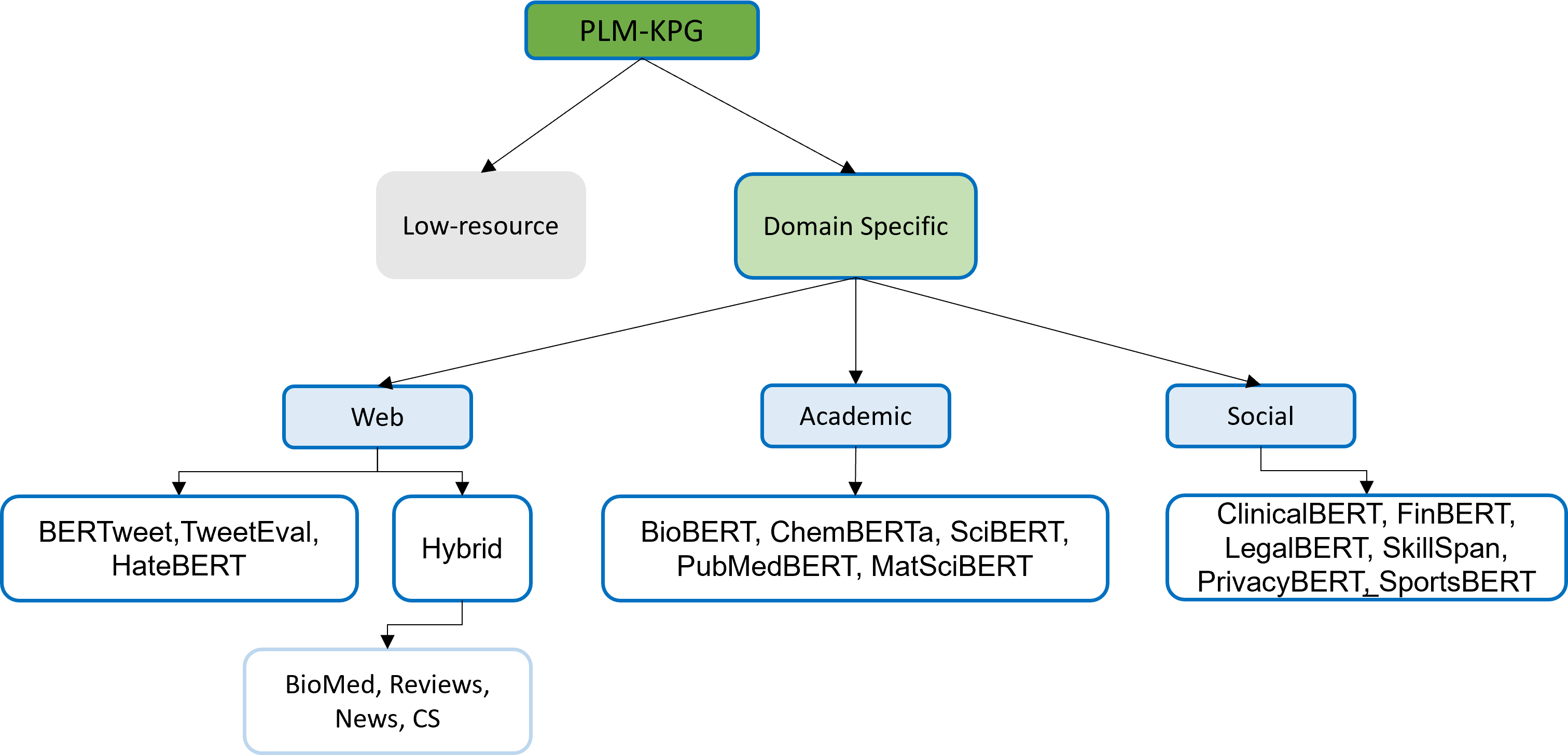}
  \caption{Categorization of Domain-specific}
  \label{DS}
\end{figure*}

There are different domain-specific PLMs that can be used for keyphrase generation. Historically, research efforts have focused on using these PLMs to achieve this particular aim. We have examined several publicly accessible PLMs that are specific to certain domains. To distinguish the level of knowledge incorporated by these PLMs, we have classified them into three types: web, academic, and social. This can be seen in the illustration provided in Figure \ref{DS}.

There is a growing body of evidence that suggests that the web domain \cite{barbieri-etal-2020-tweeteval, nguyen-etal-2020-bertweet, gururangan-etal-2020-dont, caselli-etal-2021-hatebert} is heavily influenced by empirical studies exploring the impact of incorporating PLMs into it. One such study, Twitter-roberta \cite{barbieri-etal-2020-tweeteval}, consists of seven Twitter-specific classification tasks utilizing the pre-trained language model RoBERTa \cite{liu2019roberta}. This model was trained from scratch using a Twitter corpus of 60 million tweets\footnote{584 million tokens (3.6G of uncompressed text).}, which were obtained through automatic labeling provided by Twitter\footnote{Crawled with the stream API.}. The research found that using a PLM may be sufficient, but additional training on domain-specific data can further improve performance. Micro-blogging platforms such as Twitter, where users can share real-
time information related to all kinds of topics and events. Recent evidence suggests that BERTweet \cite{nguyen-etal-2020-bertweet} investigated the differential impact of domain-specific PLM RoBERTa \cite{liu2019roberta}, and XLM-R \cite{conneau2019unsupervised} on Part-of-speech tagging, Named-entity recognition, and text classification tasks of NLP. These studies clearly indicate the effectiveness of large-scale domain-specific PLMs for English tweets. The growing interest in generating domain-specific BERT-like PLM, another work, HateBERT \cite{caselli-etal-2021-hatebert}, showed that abusive language phenomena fall along a wide spectrum, including, e.g., microaggression, stereotyping, offense, abuse, hate speech, threats, and doxxing \cite{jurgens-etal-2019-just}. It is a more robust model that obtains a higher precision score than BERT when fine-tuned to a generic abusive phenomenon \cite{caselli-etal-2020-feel}.

To determine the effects of domain-specific PLM, \cite{gururangan-etal-2020-dont, martinez2023chatgpt, wu2024leveraging} analyzed the multiple domains and showed that the adaptation of a hybrid approach incorporating four domains (Biomedical and computer science publications, news, and reviews) in PLM RoBERTa \cite{liu2019roberta}, highlights improved performance in tasks from the target domain. The study shows that domain-adaptive and task-adaptive pretraining lead to performance gains, even with limited resources. Additionally, the article provides alternative strategies for adapting to task corpora when domain-adaptive pre-trained is not feasible. The study consistently finds that multi-phase adaptive pretraining significantly improves task performance.

\begin{table*}[h]
\centering
  \caption{Datasets used for evaluation purposes are categorized based on their source types.}
  \label{tab:datasets}
  \begin{tabular}{llll}
    \toprule
    \textbf{Domain} & \textbf{Dataset} & \textbf{Created\_by} & \textbf{\#Docs}\\
    \midrule
    \textbf{News}   & DUC \cite{wan2008single} & Wan \& Xiao (2008) & 308\\
                    & KPTimes \cite{gallina2019kptimes} & Gallina et al. (2019) & 279.9K\\
                    & 500N-KPCrowd \cite{Marujo2012399} & Marujo et al.  (2012) & 500\\
    \midrule
    \textbf{Papers} & Inspec  \cite{hulth2003improved} & Hulth (2003) & 20K\\
                    & KP20K  \cite{meng2017deep} & Meng et al. (2017) & 567.8K\\
                    & KDD \cite{gollapalli2014extracting} & Gollapalli et al. (2014) &	755\\
                    & Semeval2010 \cite{kimsemeval} & Kim et al. (2010)	 & 244\\
                    & SemEval2017 \cite{augenstein2017semeval} & Augenstein et al. (2017) & 0.50K\\
                    & NUS \cite{nguyen2007keyphrase} & Nguyen and Kan (2007)    & 211\\
                    & Krapivin \cite{krapivin2009large} & Krapivin et al. (2009)   & 2.3K\\
                    & PubMed \cite{schutz2008keyphrase} & Schutz et al. (2008)     & 1.3K\\
                    & Citeulike-180 \cite{medelyan2009human} & Medelyan et al. (2009)   & 181\\
                    & TALN \cite{Boudin2013507} & Boudin, (2013)           & 521\\
                    & WWW \cite{gollapalli2014extracting} & Gollapalli (2014)        & 1.3K\\
                    & LDPK3K \cite{mahata2022ldkp} & Mahata et al. (2022)     & 96.8K\\
                    & LDPK10K \cite{mahata2022ldkp} & Mahata et al. (2022)     & 1.3M\\
    \midrule
    \textbf{Reports}& NZDL \cite{10.1145/313238.313437} & Witten et al. (1999)     & 1.8K\\
    \midrule
    \textbf{Web}    & Blogs \cite{10.1145/1526709.1526798} & Grineva et al. (2009)    & 252\\
                    & OpenKP \cite{xiong2019open} & Xiong et al. (2019) & 147.20K\\
    \midrule
    \textbf{QA}     & StackExchange \cite{wang-etal-2019-topic-aware} & Wang et al. (2019)       & 49.4K\\
    \midrule
    \textbf{Tweets} & Twitter \cite{zhang-etal-2016-keyphrase} & Zhang et al. (2016)      & 112.5K\\
                    & Text-Image Tweets \cite{wang-etal-2020-cross-media} & Wang, Li et al. (2020)   & 53.7K\\

  \bottomrule
\end{tabular}
\end{table*}

The second subcategory is academic, A number of existing works adopted PLMs \cite{10.1093/bioinformatics/btz682, chithrananda2020chemberta, beltagy-etal-2019-scibert, 10.1145/3458754, DBLP:journals/corr/abs-2109-15290, salaun2024europa}. \cite{10.1093/bioinformatics/btz682} This research offers fresh perspectives on biomedical text mining. After realizing that BERT \cite{devlin2018bert}, a well-known PLM, excels at general language tasks, it frequently provides wrong responses to straightforward biomedical queries. In order to enhance BERT's performance on biomedical text mining tasks such as question answering (QA), which highly relay on quality text generation \cite{yoon2019pre}, researchers pre-trained BERT on sizable biomedical corpora. In \cite{chithrananda2020chemberta} ChemBERTa, a large-scale self-supervised pretrained model for molecular property prediction using transformers. The authors assess the effectiveness of transformers in learning molecular representations and predicting properties. They demonstrate competitive performance on MoleculeNet and use functional attention-based visualization techniques. SciBERT \cite{beltagy-etal-2019-scibert} is a sophisticated model that underwent rigorous training in a comprehensive corpus of 1.14 million biomedical and computer science papers from the Semantic Scholar database. This model was trained using unsupervised pre-training in the academic domain. The model's performance was evaluated in \cite{wu2022pre} through a series of tasks and datasets from scientific fields, which confirmed its position as the leading scientific keyphrase generation model in the industry. Recent developments in natural language processing have led to the creation of domain-specific language models like MatSciBERT \cite{gupta2021matscibert} for the materials domain, which outperforms general-domain models in various tasks. Similarly, in the biomedical domain \cite{10.1145/3458754}, starting from scratch with domain-specific pre-training yields significant improvements over using general domain language models, as demonstrated by the BLURB benchmark. These findings emphasizImportancertance of domain-specific pre-training for keyphrase generation in specialized fields.

Finlay, in this survey, we explore the application of domain-specific pre-trained language models for keyphrase generation in the social domain. Several studies \cite{alsentzer-etal-2019-publicly, liu2021finbert, chalkidis-etal-2020-legal, zhang-etal-2022-skillspan, srinath-etal-2021-privacy, 3l1nnz0z39, 10.1007/978-981-99-7019-3_29} have addressed the need for specialized language models in various fields. \cite{alsentzer-etal-2019-publicly} and \cite{liu2021finbert} focus on clinical and financial domains, respectively, and demonstrate that domain-specific pre-training leads to improved performance on related NLP tasks. In \cite{chalkidis-etal-2020-legal, cheong2024not}, the study explores the adaptation of BERT and LLMs in the legal domain and proposes different strategies for better results. \cite{zhang-etal-2022-skillspan} introduces a new dataset and shows the effectiveness of domain-adapted models for skill extraction. \cite{srinath-etal-2021-privacy} introduces a large corpus of privacy policies to facilitate the creation of privacy-related models. Lastly, \cite{3l1nnz0z39} presents SportsBERT, a domain-specific BERT model for the sports domain, emphasizing the benefits of training models tailored to specific domains. These findings collectively emphasizImportancertance of domain-specific pre-training for social keyphrase generation tasks in specialized fields.

\section{Models Comparison}
\label{modelcomparison}
To rigorously assess the efficacy of various keyphrase extraction and generation models, we meticulously selected one emblematic model from each methodological category for comparative analysis listed in Table \ref{Comparison}. Our selection criteria were based on each model's prominence in the field, innovative approach, and reported performance in prior studies. To ensure a comprehensive evaluation that accounts for varying document complexities, we employed two distinct datasets: one comprising short documents, which presents a challenge in distilling key content, and another consisting of long documents, demanding effective processing of sizeable information. The detailed characteristics of these datasets, including their average document length, domain diversity, and dataset created\_by, are delineated in Table \ref{tab:datasets}. This selection strategy enables us to provide a balanced and insightful comparison that sheds light on the strengths and limitations of each model across different textual contexts.

\subsection{Attention Adaptation}
\label{am}
In this section, we will explore various advanced attention mechanisms that have the potential to enhance the performance of KPE and KPG tasks significantly. These mechanisms offer unique advantages in handling long and short sequences and improving computational efficiency. By leveraging these cutting-edge techniques, researchers can develop more efficient and accurate models, ultimately advancing the field of NLP and enriching applications in keyphrase extraction. Each mechanism will be discussed, highlighting its specific benefits and contributions to the task at hand.

\begin{enumerate}
\item \textbf{Sparse Attention Mechanisms}: Sparse attention mechanisms focus on specific parts of the input sequence, reducing computational complexity by ignoring less relevant parts. This efficiency makes sparse attention particularly useful for handling long documents in keyphrase extraction, ensuring that the model can process longer texts without being overwhelmed by the entire input.\cite{roy2021efficient}.
    
\item \textbf{Blockwise Attention}: Blockwise attention divides the input sequence into blocks and computes attention within each block, rather than across the entire sequence. This method reduces computational load and memory usage, allowing the model to process longer texts more effectively. For keyphrase extraction, blockwise attention ensures that attention is localized and computationally feasible, making it easier to capture context-specific keyphrases from large documents \cite{liu2023blockwise}.

\item \textbf{Linformer}: Linformer approximates the full attention mechanism by projecting the input sequence into a lower-dimensional space, significantly reducing the computational complexity. This efficiency makes Linformer suitable for tasks requiring attention over long sequences, such as keyphrase extraction, where it can maintain performance while being computationally more efficient \cite{wang2020linformer}.

\item \textbf{Reformer}: Reformer uses locality-sensitive hashing to reduce the quadratic complexity of the attention mechanism to logarithmic complexity, enabling efficient processing of very long sequences. For keyphrase generation, Reformer allows the model to handle extensive documents while maintaining the ability to focus on relevant parts of the text, thus improving the extraction of meaningful keyphrases \cite{kitaev2020reformer}.

\item \textbf{Ring Attention}: Ring attention introduces a mechanism where attention is computed in a circular manner, focusing on both nearby and distant tokens in a structured way. This approach can enhance the model's ability to capture long-range dependencies, which is crucial for identifying keyphrases that are contextually relevant but dispersed throughout the document \cite{liu2023ring}.

\item \textbf{Longformer}: Longformer extends the Transformer model by incorporating dilated sliding window attention, allowing it to handle longer sequences with reduced computational cost. This mechanism is particularly beneficial for keyphrase extraction as it enables the model to consider extensive contexts without sacrificing efficiency, ensuring comprehensive extraction of keyphrases from lengthy texts \cite{beltagy2020longformer}.

\item \textbf{Adaptive Attention Span}: Adaptive attention span dynamically adjusts the attention span for different tokens based on their importance, optimizing computational resources. This adaptability is crucial for keyphrase extraction, where the relevance of information can vary significantly throughout the document. By focusing more on important sections, the model can extract more pertinent keyphrases efficiently \cite{sukhbaatar2019adaptive}.
\end{enumerate}

\begin{table*}[h!]
\centering
\caption{The comprehensive comparison of model precision (P), recall (R), and F-score (F1) across various thresholds, namely @5, @10, and @15, using two benchmark datasets. The variable \textit{N} represents the number extracted from a single document by the models.}
\begin{tabular}{lccccccc}
\hline
\multirow{2}{*}{\textbf{N}} &
  \multirow{2}{*}{\textbf{Method}} &
  \multicolumn{3}{c}{\textbf{Inspec}} &
  \multicolumn{3}{c}{\textbf{SemEval2010}} \\ \cline{3-8} 
 &
   &
  \textbf{P} &
  \textbf{R} &
  \textbf{F1} &
  \textbf{P} &
  \textbf{R} &
  \textbf{F1} \\
  \hline
\multirow{4}{*}{\textbf{5}}  & \textbf{AttentionRank} & 41.19 & 17.38 & 24.45 & 22.27 & 7.66 & 11.39 \\                               & \textbf{TextRank}  & 36.16 & 18.40 & 24.39 & 36.63 & 10.59 & 16.43 \\
                           & \textbf{MDERank}       & 32.25 & 16.52 & 27.85 & 32.52 & 11.67 & 13.05 \\
                           & \textbf{SIFRank}       & 44.00 & 18.40 & 25.95 & 11.44 & 3.83 & 5.74 \\
                           \hline
\multirow{4}{*}{\textbf{10}} & \textbf{AttentionRank} & 37.17 & 28.32 & 32.15 & 18.65 & 12.72 & 15.12 \\
                           & \textbf{TextRank}  & 33.44 & 33.71 & 33.58 & 35.25 & 20.38 & 25.83 \\
                           & \textbf{MDERank}       & 35.41 & 32.87 & 34.36 & 36.40 & 31.67 & 18.27 \\
                           & \textbf{SIFRank}       & 37.35 & 29.43 & 32.92 & 7.82 & 5.18 & 6.23 \\ 
                           \hline
\multirow{4}{*}{\textbf{15}} & \textbf{AttentionRank} & 34.58 & 34.40 & 34.49 & 16.51 & 16.82 & 16.66 \\
                           & \textbf{TextRank}  & 30.09 & 44.14 & 35.78 & 32.86 & 28.46 & 30.50 \\            
                           & \textbf{MDERank}       & 33.45 & 40.42 & 36.40 & 32.48 & 31.59 & 20.35 \\
                           & \textbf{SIFRank}       & 32.51 & 35.73 & 34.04 & 6.20 & 6.11 & 6.15 \\
\hline
\end{tabular}
\label{Comparison}
\end{table*}

\begin{table*}[h!]
    \centering
    \begin{tabular}{llllllll}
    \toprule
      \textbf{Dataset} & \textbf{Keyphrases/Doc} & \textbf{Abstract/Doc} & \textbf{Tokens/Doc} & \textbf{Keyphrase Length} & \textbf{Abstract Length} & \textbf{Docs} \\
      \midrule
       KP20k & 3.24 & 2.84 & 179.02 & 1.85 & 2.55 & 570,802 \\
       SemEval & 6.25 & 8.41 & 245.89 & 2.08 & 2.61 & 100 \\
       Krapivin & 3.26 & 2.59 & 189.32 & 2.16 & 2.29 & 400 \\
       NUS & 6.34 & 5.31 & 230.13 & 1.95 & 2.56 & 211 \\
       Inspec & 7.23 & 2.59 & 134.10 & 2.44 & 2.72 & 500 \\
    \bottomrule
    \end{tabular}
    \caption{Statistical features of five datasets.}
    \label{tab:statistics}
\end{table*}

\section{Dataset and SOTA PLM-KP Baselines methods}
\label{dataset}

\subsection{Datasets}
Datasets from a variety of text sources, including news items, abstracts of journal papers, and full-text scientific publications, have been used to evaluate KP systems. Table \ref{tab:datasets} lists some of the most popular KP datasets, organized by domain type. We list the name (Dataset), creator(s) (Created By), and number of text documents (\#Docs) for each dataset. They could be separated into news, research papers, reports, web pages, QA, and tweets, according to domains. These datasets are primarily in English.

The most commonly used dataset for KP is KP20k, which contains articles on computer science from various online libraries. However, due to their small size, these datasets are unsuitable for industrial applications. To advance the field of KP, it is crucial to create many high-quality multilingual datasets. The available resources can be found on Github. 
\footnote{\url{https://github.com/snkim/AutomaticKeyphraseExtraction}},
\footnote{\url{https://github.com/zelandiya/keyword-extraction-datasets}}.

Table \ref{tab:statistics} displays the statistical information related to the most widely used datasets used to predict keyphrases within the natural language processing domain. These details encompass magnitude levels, the count of present and absent keyphrases per document, the number of tokens contained within each document, the length of both present and absent keyphrases, and the total number of documents. These statistics hold significant value for professionals seeking to remain informed of the latest trends and patterns within this field of research.

\subsection{SOTA PLM-KP Baselines methods}
Within this particular subsection, we delve into the most up-to-date models for unsupervised keyphrase extraction, that leverage pre-trained language models as their foundation listed in Table \ref{tab:sotamodels}. Based on the transformer architecture and trained using the AdamW optimization algorithm, these models serve as the benchmark for evaluating the effectiveness of advanced keyphrase extraction methods, setting a robust standard for future research and development in the domain of pre-trained language models for keyphrase extraction.

These models represent the pinnacle of current advancements in the field, showcasing their remarkable performance in accurately extracting keyphrases from diverse textual datasets. The SIFRank model \cite{sun2020sifrank} boasting 1.3 billion parameters and trained on a corpus of 1.5 billion words, demonstrates exceptional accuracy at 84.7\%, coupled with impressive speed, being the smallest model with 1.3 billion parameters and JointGL-large \cite{liang2021unsupervised} being the largest model with 340 million parameters. The number of parameters of a model is indicative of its complexity and capability to learn more sophisticated patterns in data. However, the use of larger models can pose challenges in terms of resource requirements. It is important to note that while the accuracy of models tends to improve with an increase in parameters, training time and speed also increase accordingly. All models are trained on a vast corpus of text to learn patterns of word usage. To determine the accuracy of the models in extracting keyphrases, a benchmark dataset of keyphrases is used. The speed of the models is based on the time it takes to extract keyphrases from a document.

The architecture of a language model refers to its structure, including the number and type of layers and their connections. Architecture can affect the accuracy, speed, and efficiency of the model. The training algorithm employed also plays a crucial role, including the optimization algorithm, the learning rate, and the applied regularization techniques. The choice of training algorithm can influence the accuracy, speed, and stability of the model. In assessing language models, various metrics such as accuracy, precision, recall, and the F1 score are utilized to determine their strengths and weaknesses. These metrics provide a comprehensive evaluation of the model's performance and assist in identifying areas that require improvement.

\begin{table*}[h!]
\centering
\begin{tabular}{lllllll}
\toprule
\textbf{ Model} & \textbf{Param} & \textbf{Tra. data} & \textbf{Acc.} & \textbf{Speed} &\textbf{ Model arch.} & \textbf{Training alg.} \\
\midrule
SIFRank \cite{sun2020sifrank} & 1.3B & 1.5B words & 84.7\% & Fast & Transformer & AdamW \\
JointGL \cite{liang2021unsupervised} & 117M & 400B words & 87.8\% & Fast & Transformer & AdamW \\
JointGL \cite{liang2021unsupervised} & 340M & 400B words & 90.7\% & Slow & Transformer & AdamW \\
MDERank \cite{zhang2021mderank} & 137M & 400B words & 89.6\% & Fast & Transformer & AdamW \\
\bottomrule
\end{tabular}
\caption{State-of-the-art PLM-KP models.}
\label{tab:sotamodels}
\end{table*}

\section{Exploring the Landscape of PLM-KP: Applications, Challenges, and Future Research Directions}
\label{AppChalFRD}
Pre-trained language models (PLMs) are now an essential tool in NLP and have significantly impacted keyphrase extraction and generation. Leveraging vast amounts of textual, visual, and audio data, PLMs can identify important information and understand the language context. This capability allows them to capture intricate patterns, making PLMs particularly useful in keyphrase extraction, where relevance and precision are critical. Researchers and practitioners can extract meaningful keyphrases from diverse textual sources, improving content organization, retrieval of information, and user experience. This section explores the various applications of PLMs and their significant role in advancing the field of keyphrase extraction and generation, as well as highlighting the challenges and opportunities that lie ahead.

\subsection{Applications of PLM-KP}
\label{applications}
Advances in language models have revolutionized how systems understand textual, visual, and audio data, enabling more precise and efficient keyphrase prediction. Their ability to recognize subtle nuances has empowered researchers and industry professionals to extract valuable insights from extensive datasets. This subsection explores their diverse applications and highlights their impressive capacity for extracting precise keyphrases, demonstrating their transformative impact on information retrieval and content enrichment.

\subsubsection{Academic Publishing}
Pre-trained language models (PLMs) have significantly advanced academic publishing by automating the extraction and generation of keyphrases. This capability is crucial for indexing and metadata generation, enhancing research papers' discoverability in scholarly databases. For instance, PLMs like BERT and GPT-3 analyze extensive research texts to extract meaningful keyphrases, aiding in the creation of abstracts and keyword lists. This automated process not only saves time but also ensures consistency and accuracy in metadata generation, making academic documents more searchable and retrievable \cite{zhang2022enhancing}. Additionally, PLMs are used to identify emerging research trends and significant contributions within specific fields, facilitating literature reviews by summarizing vast amounts of information into concise keyphrases, helping researchers quickly grasp the latest developments and identify knowledge gaps \cite{gollapalli2014extracting}. Platforms like SCISpace \footnote{SCISpace \url{https://scispace.com/}} showcase real-world applications of these models in enhancing academic research environments.

\subsubsection{Content Management and SEO}
In content management and search engine optimization (SEO), pre-trained language models (PLMs) are indispensable for enhancing the visibility and organization of web content. These models can extract keyphrases from digital content, including blogs, articles, and product descriptions, identifying the most pertinent terms to boost search engine rankings. Integration of models like Google Gemini \footnote{Google Gemini \url{https://gemini.google.com/}} and Perplexity AI \footnote{Perplexity \url{https://www.perplexity.ai/}} improves the user experience by understanding and responding to user intent with precise information. PLMs analyze text to generate keyphrases that align with common search queries, optimizing content to attract organic traffic. Their ability to comprehend context and semantics ensures that the extracted keyphrases are relevant and strategically positioned to enhance SEO efforts effectively \cite{rose2010automatic}.

\subsubsection{Business Intelligence}
In business intelligence, pre-trained language models are leveraged to extract and generate keyphrases from vast amounts of textual data, enabling companies to gain actionable insights from market research reports, customer feedback, and internal documents. PLMs can process diverse business texts to identify key trends, competitive analyses, and consumer sentiments, summarizing them into concise keyphrases that inform strategic decision-making \cite{gollapalli2014extracting}. This helps businesses stay ahead of market trends, understand customer needs, and improve product offerings based on extracted insights. Furthermore, analyzing customer feedback and reviews allows businesses to enhance customer satisfaction and loyalty by identifying common issues, preferences, and sentiments, facilitating a data-driven approach to product development and marketing strategies.

\subsubsection{Healthcare}
The healthcare industry benefits immensely from employing pre-trained language models (PLMs) for keyphrase extraction and generation. PLMs can analyze clinical notes, patient records, and biomedical literature to extract keyphrases summarizing critical information, assisting healthcare professionals in making informed decisions swiftly. Models like BioBERT are specifically trained on biomedical texts, enabling accurate extraction of keyphrases highlighting important patient details and medical findings, thus enhancing patient care by providing rapid access to essential information \cite{lee2020biobert, alsentzer2019publicly}. In biomedical research, PLMs aid in extracting keyphrases from scientific articles, helping researchers identify significant trends and discoveries, making literature reviews more efficient and aiding in the discovery of new treatments and therapies \cite{lee2020biobert}.

\subsubsection{Legal Compliance}
In the legal sector, pre-trained language models (PLMs) are used to extract keyphrases from legal documents, contracts, and regulatory filings, streamlining the review process and ensuring compliance with industry standards. PLMs can analyze complex legal texts to identify key terms and provisions, facilitating navigation and understanding of extensive documents \cite{sleimi2021automated}. This automated extraction reduces the time and effort required for manual review, allowing legal teams to focus on strategic tasks, improving overall efficiency. Additionally, PLMs help maintain regulatory compliance by extracting keyphrases that highlight critical regulatory requirements and changes, which is particularly valuable for heavily regulated industries like finance and healthcare \cite{gollapalli2014extracting}.

\subsubsection{News and Media}
Pre-trained language models have transformed the news and media industry by improving the efficiency and accuracy of news summarization and trend analysis. These models can extract key phrases from news articles to create concise summaries, making it easier for readers to quickly grasp the main points of a story \cite{deka2022improved}. This capability is especially beneficial for news aggregators and media monitoring services, which need to process large volumes of content and provide timely updates. By automating the summarization process, PLMs ensure that readers have access to relevant and up-to-date information without being overwhelmed by the sheer volume of news. Additionally, PLMs play a crucial role in identifying trending topics by analyzing large sets of news articles, helping journalists and media analysts stay informed about current trends and public sentiments.

\subsection{Challenges of PLM-KP}
\label{challenges}
In general, the prediction of keyphrases has gained
significant interest from both academic and industrial
sectors. Nevertheless, it is still considered a challenging
task due to various following facets:
\begin{enumerate}
\item \textbf{Absent Keyphrases}: Generating high-quality absent keyphrases, which are not explicitly mentioned in the text but are contextually relevant, remains a significant challenge. Current models often struggle with predicting these phrases accurately because they require a deep understanding of the context and the ability to infer missing information. Improving this aspect involves developing more sophisticated models that can better understand and infer context, potentially by incorporating external knowledge bases or enhanced training techniques. Future research should focus on creating algorithms that can more effectively identify and generate absent keyphrases, as this would significantly improve the quality and relevance of keyphrase generation.

\item \textbf{External Knowledge Integration}: Effectively integrating external knowledge bases or pre-trained models is essential for optimizing keyphrase prediction. Current models may not fully leverage external data sources such as encyclopedias, databases, or domain-specific corpora, which can provide additional context and improve the accuracy of generated keyphrases. Incorporating external knowledge can help models understand nuances and context that are not immediately apparent in the input text, leading to more precise and relevant keyphrases. This integration poses a technical challenge, requiring advanced methods to seamlessly combine and utilize these diverse data sources.

\item \textbf{Multi-modal Data}: The increasing prevalence of multimedia content necessitates a multi-modal approach to keyphrase prediction. Traditional text-based models struggle to handle data that includes audio, video, and images, which are becoming more common in digital communication. Developing models that can process and integrate these different types of data to generate coherent and contextually appropriate keyphrases is a complex challenge. This requires advancements in multi-modal learning and the ability to align and synthesize information from various modalities effectively.

\item \textbf{Domain Adaptation}: Models trained on domain-specific data often perform poorly when applied to other domains due to differences in terminology, context, and writing styles. Effective transfer mechanisms are needed to adapt these models to diverse real-world settings without significant loss in performance. This involves techniques such as domain adaptation, transfer learning, and fine-tuning, which can help models generalize better across different types of data. Developing robust methods for domain adaptation will enable the application of keyphrase generation models in a wider range of industries and contexts.

\item \textbf{Semantic Evaluation}: Current evaluation metrics for keyphrase generation typically rely on exact matches, which do not account for semantic similarity. This can lead to underestimation of a model's performance if the generated keyphrases are semantically similar but not exact matches to the reference keyphrases. Implementing evaluation metrics that consider semantic similarity will provide a more accurate assessment of the quality of generated keyphrases. This involves using techniques such as embedding-based similarity measures and human-in-the-loop evaluations to capture the true relevance and meaning of keyphrases \cite{boudin2021keyphrase}.

\item \textbf{Separate Generation of Keyphrases}: Considering separate generation processes for present and absent keyphrases can improve prediction accuracy. Recent research suggests that treating these tasks separately allows for more targeted model training and better handling of each type's unique challenges \cite{wang2020minilm}. Additionally, emulating human reading and refinement techniques, such as extracting the general idea before focusing on details, can enhance the prediction process. This coarse-to-fine approach mimics how humans process information and can lead to more accurate keyphrase generation.

\item \textbf{Utilizing ChatGPT}: Investigating how to best leverage ChatGPT for keyphrase prediction is crucial to maximize its potential. While ChatGPT has shown impressive proficiency in various NLP tasks, fine-tuning it specifically for keyphrase generation and understanding the best prompt designs and use cases can significantly enhance its performance. Exploring these aspects will help in utilizing ChatGPT's capabilities more effectively for keyphrase extraction and generation. One study took the first step towards this challenge \cite{song2023chatgpt}.

\item \textbf{Multi-modal Integration}: Further research is needed to effectively integrate multi-modal signals for keyphrase prediction. Combining text, images, and other attributes presents a significant challenge but also offers an opportunity to enrich the generated keyphrases with more context and relevance. This integration requires advanced techniques in multi-modal learning and the ability to process and fuse different types of data seamlessly.

\item \textbf{Large-scale Datasets}: Expanding the size of keyphrase prediction datasets is necessary to enhance model applicability in industrial contexts. Most current datasets are relatively small and domain-specific, limiting the generalizability of the models. Larger and more diverse datasets will provide better training material, allowing models to learn a broader range of contexts and applications, thereby improving their performance in real-world scenarios.
\end{enumerate}

\subsection{Future Research Directions}
\label{future}
\begin{enumerate}
\item \textbf{Diversity:} Ensuring the diversity of generated keyphrases and extracted is a critical research direction to enhance the applicability and usefulness of keyphrase extraction models. Diversity in keyphrases ensures that various aspects of the content are covered, reducing redundancy and providing a comprehensive summary of the document. To address this, future research should focus on developing techniques that promote the generation of diverse keyphrases, such as using diversification algorithms, penalizing repetitive outputs, and incorporating coverage mechanisms that ensure a wide range of topics and ideas are included. Evaluating diversity through metrics that measure uniqueness and coverage will also be essential to advance this area \cite{xie2022wr}.

\item \textbf{Attention Mechanism:} Exploring advanced attention mechanisms can significantly enhance the performance of keyphrase extraction models. Techniques such as sparse attention, blockwise attention, Linformer, Reformer, Ring Attention, Longformer, and adaptive attention spans offer different ways to handle large texts and focus on relevant parts of the input. Investigating these mechanisms can lead to more efficient and accurate keyphrase generation by improving the model's ability to capture and process important information within the text.

\item \textbf{Evaluation Metric:} Traditional evaluation metrics for keyphrase generation, such as those based on exact matches (F1-based metrics), often fail to capture partial matches or semantic similarities between predicted and gold keyphrases \cite{ye2021one2set}. For instance, "keyphrase generation model" and "keyphrase generation system" may be semantically similar but would not be recognized as such by exact match metrics. Therefore, developing semantic-based evaluation metrics that account for meaning and context will provide a more accurate measure of model performance. Additionally, incorporating human evaluation can offer deeper insights into the effectiveness and practicality of keyphrase generation models.

\item \textbf{Prompt Designing:} The effectiveness of ChatGPT in generating keyphrases significantly depends on the design of the prompts used. While we have improved the prompts based on OpenAI's guidelines, they are not necessarily optimal. Developing more refined and context-specific prompts is crucial to fully exploit ChatGPT's capabilities for keyphrase generation. Crafting prompts that precisely guide the model to focus on relevant aspects of the text can enhance the quality and relevance of the generated keyphrases, making this an essential area for future research.
    
\item \textbf{Hyper-Parameter:} In practical applications, users typically do not focus on adjusting the hyper-parameters of ChatGPT. However, the settings of these parameters, which usually require in-depth knowledge about the model, can greatly influence its performance. Future research should systematically study the impact of various hyper-parameters on keyphrase generation to optimize the model's performance and provide guidelines for users without requiring extensive technical expertise.

\item \textbf{Few-Shot Prompting:} The rise of in-context learning with large language models has introduced a new paradigm where models make predictions based on a few examples provided in the context. This approach, known as few-shot prompting, has shown promising results in various NLP tasks \cite{dong2022survey, wei2022emergent}. Exploring these strategies in future research should focus on applying few-shot learning techniques to enhance the performance of ChatGPT and similar models in keyphrase generation. This can significantly reduce the amount of labeled data needed and improve model adaptability.
\end{enumerate}

\section{Conclusion}
\label{concl}
This study comprehensively examines the evolving landscape of Pre-trained Language Models (PLMs) in the context of Keyphrase Prediction (KP). The introduction underscores the significance of KP in summarizing document content, and the rapid advancements in Natural Language Processing have paved the way for more efficient KP models employing deep learning techniques. The taxonomy introduced for Pre-trained Language Model Keyphrase Extraction (PLM-KPE) and Pretrained Language Model Keyphrase Generation (PLM-KPG) captures the core focus of these tasks, highlighting their distinct aspects. The main contributions of this research lie in
delineating the latest developments in neural network-based keyphrase prediction, elucidating their applications in PLMKPE and PLM-KPG, and recognizing their unique perspectives. The discussion showcases an in-depth categorization of PLMs and explores specific strategies within PLM-KPE and PLM-KPG domains. Incorporating self-supervised learning methods and the contributions of models such as BERT, GPT, and T5 underscore the dynamic landscape of PLMKP. The illustrative example highlights the dual categories of keyphrases, consistently present and absent, underscoring the nuanced nature of keyphrase prediction. This paper bridges the gap between PLM advancements and their application to KP, laying the foundation for future research avenues and advancements in the field of NLP keyphrase prediction.
\section*{Acknowledgments}
This work was supported by Institute of Information \& communications Technology Planning \& Evaluation (IITP) grant funded by the Korea government(MSIT) (No.RS-2022-00155911, Artificial Intelligence Convergence Innovation Human Resources Development (Kyung Hee University))

\section*{Conflict of interest}
The authors declare no conflict of interest.

\bibliographystyle{elsarticle-num}
\bibliography{ref}

\end{document}